\documentclass[10pt,twocolumn,letterpaper]{article}

\usepackage{iccv}

\usepackage{times}
\usepackage{epsfig}
\usepackage{graphicx}
\usepackage{amsmath}
\usepackage{amssymb}
\usepackage{multirow}
\usepackage{bbm}
\usepackage{algorithmicx}
\usepackage[ruled,linesnumbered]{algorithm2e}
\usepackage[table,x11names]{xcolor}
\usepackage{amssymb}
\usepackage{pifont}
\usepackage{diagbox}
\usepackage[accsupp]{axessibility}

\usepackage[pagebackref=true,breaklinks=true,letterpaper=true,colorlinks,bookmarks=false]{hyperref}

\iccvfinalcopy
\def\iccvPaperID{6622} 

\ificcvfinal\fi

\newcommand{\methodname}{MetaUVFS}

\begin{document}

\title{Unsupervised Few-Shot Action Recognition via Action-Appearance Aligned Meta-Adaptation}

\author{Jay Patravali*\\
Microsoft\\
{\tt\small t-jpatra@microsoft.com}
\and
Gaurav Mittal*\\
Microsoft\\
{\tt\small gaurav.mittal@microsoft.com}

\and
Ye Yu\\
Microsoft\\
{\tt\small yu.ye@microsoft.com}

\and
Fuxin Li\\
Oregon State University\\
{\tt\small lif@oregonstate.edu}

\and
Mei Chen\\
Microsoft\\
{\tt\small mei.chen@microsoft.com}
}

\author{Jay Patravali$^\star$$^\ddag$ ~~~~ Gaurav Mittal$^\star$$^\dag$ ~~~~ Ye Yu$^\dag$ ~~~~ Fuxin Li$^\ddag$ ~~~~  Mei Chen$^\dag$\\
$^\dagger$Microsoft ~~~~~~~~~~~~~~~~~~~~~~~~ $^\ddag$Oregon State University\\
{\tt\small \{gaurav.mittal, yu.ye, mei.chen\}@microsoft.com} ~~~
{\tt\small \{patravaj,lif\}@oregonstate.edu}
}

\maketitle
\thispagestyle{empty}

\begin{abstract}
We present MetaUVFS as the first \textbf{U}nsupervised \textbf{Meta}-learning algorithm for \textbf{V}ideo \textbf{F}ew-\textbf{S}hot action recognition. MetaUVFS leverages over 550K unlabeled videos to train a two-stream 2D and 3D CNN architecture via contrastive learning to capture the appearance-specific spatial and action-specific spatio-temporal video features respectively. MetaUVFS comprises a novel Action-Appearance Aligned Meta-adaptation~(A3M) module that learns to focus on the action-oriented video features in relation to the appearance features via explicit few-shot episodic meta-learning over unsupervised hard-mined episodes. Our action-appearance alignment and explicit few-shot learner conditions the unsupervised training to mimic the downstream few-shot task, enabling MetaUVFS to significantly outperform all state-of-the-art unsupervised methods on few-shot benchmarks. Moreover, unlike previous few-shot action recognition methods that are supervised, MetaUVFS needs neither base-class labels nor a supervised pretrained backbone. Thus, we need to train MetaUVFS just once to perform competitively or sometimes even outperform state-of-the-art supervised methods on popular HMDB51, UCF101, and Kinetics100 few-shot datasets.
\vspace{-1em}
\end{abstract}



\section{Introduction}
\let\thefootnote\relax\footnote{$^\star$ Authors with equal contribution.}
\let\thefootnote\relax\footnote{Work done when Jay Patravali was a full-time intern at Microsoft.}
\begin{figure}[t]
\begin{center}
    \includegraphics[width=\linewidth]{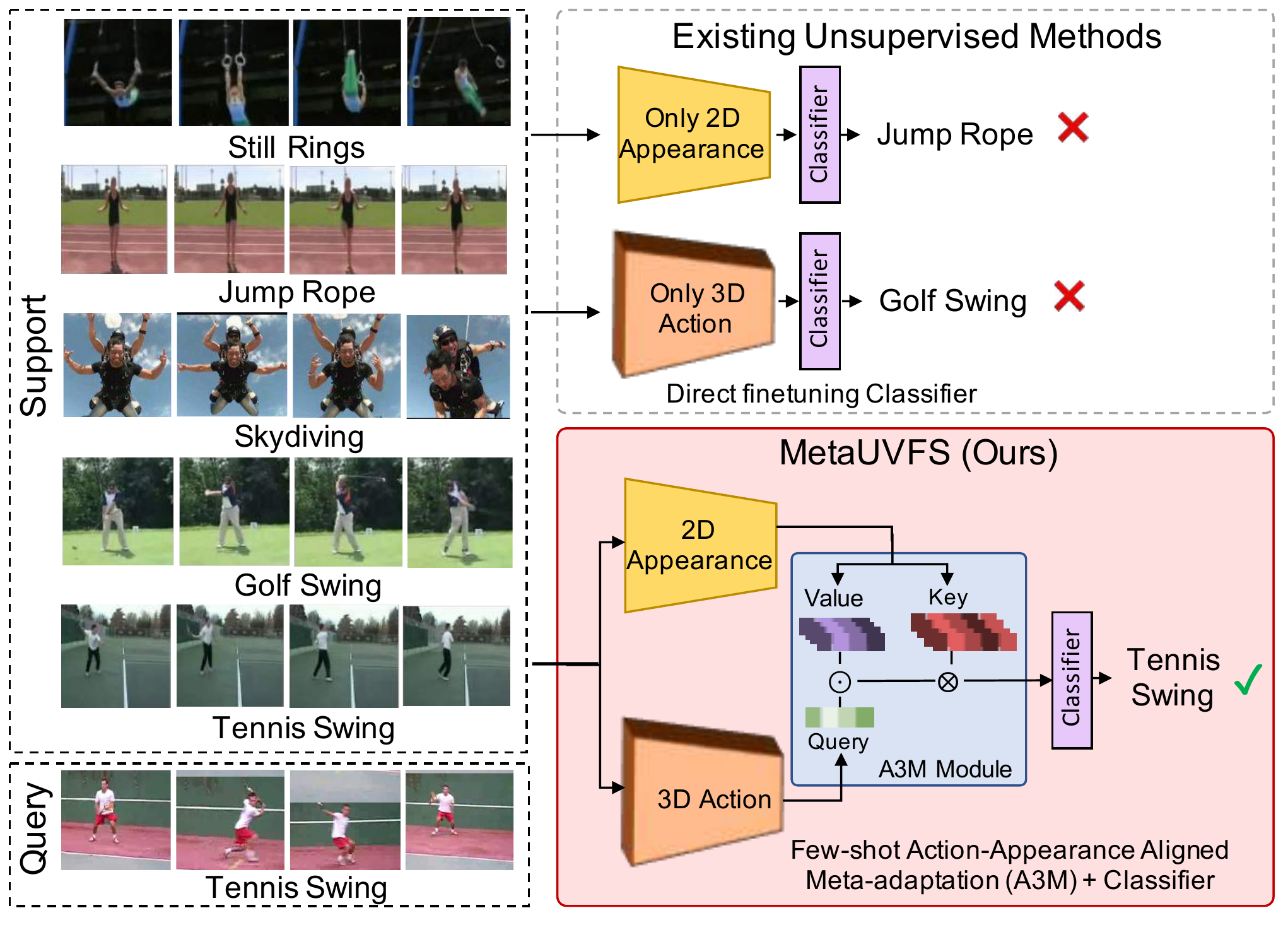}

\end{center}
\vspace{-1em}
   \caption{The above example shows a 1-support 5-way few-shot video action recognition task to classify a query sample of novel \textit{Tennis Swing} class.  Using only appearance with a 2D CNN incorrectly predicts \textit{Jump Rope} as it relies on only frame-level spatial cues. Using only action with a 3D CNN incorrectly predicts \textit{Golf Swing} as it matches based on just the swinging action without paying attention to the spatial cues. Whereas \methodname~predicts the correct class via the Action-Appearance Aligned Meta-adaptation~(A3M) module that learns to align and relate the action with the appearance attributes via few-shot meta-learning. All three methods are trained using unlabeled videos.
}
\label{fig:teaser}
\vspace{-1em}
\end{figure}

Few-shot learning~\cite{koch2015siamese, sung2018learning, vinyals2016matching, gidaris2018dynamic, snell2017prototypical, ravi2016optimization, koch2015siamese, finn2017model, chen2019closer, hariharan2017low, wang2018low} has emerged as a school of approaches that train a model to transfer-learn or adapt quickly on novel, often out-of-domain, classes using as few labeled samples as possible to mitigate the lack of large-scale supervision for these novel classes. Few-shot learning is highly relevant for videos because collecting large-scale labeled video data is extra challenging with the additional temporal dimension. There has been work utilizing both 2D and 3D CNNs~\cite{zhu2018compound,Cao_2020_CVPR, fu2020depth, xian2020generalized,cao2020few, zhang2020few} to achieve strong results on few-shot action recognition in videos. However, these are supervised approaches and require large amounts of labeled base-class data and/or large-scale supervised pretrained backbones~\cite{Cao_2020_CVPR, cao2020few, fu2020depth, xian2020generalized} that are not only prohibitively expensive to scale but also oftentimes unattainable. Meanwhile, there is virtually infinite unlabeled video data at our disposal through the rise of multi-media social networking. This motivates us to address the question, \textit{``Can we develop models for video action recognition that perform competitively on few-shot benchmarks without the use of either base-class labels or any external supervision?’’}

Existing unsupervised video representation learning methods~\cite{qian2020spatiotemporal, tao2020self,  Han20CoCLR}  provide task-agnostic representations that apply to various downstream tasks. However, as we show in later sections, these methods are not specialized for the few-shot learning task with novel classes and therefore perform sub-optimally on them. 

To this end, we propose \methodname~as the first method for unsupervised meta-learning for few-shot video action recognition. \methodname~leverages large-scale~(over half a million) unlabeled video data to learn video representations via contrastive learning and then trains an explicit few-shot meta-learner using episodes that are hard-mined over the learned representations. The episodic meta-learning helps mimic the episodic few-shot meta-testing during the training phase. This imposes a downstream task-specific prior on the learned video representations and reduces the knowledge gap between training and testing. 

We introduce an unsupervised two-stream action-appearance network in \methodname~to learn fine-grained spatio-temporal 3D features over video segments via an action stream and spatial 2D features over video frames via an appearance stream. Direct finetuning of either feature alone can be sub-optimal in a challenging few-shot scenario as illustrated in Fig.~\ref{fig:teaser}. Instead, we design an Action-Appearance Aligned Meta-adaptation module~(A3M) in the few-shot meta-learner of \methodname~that combines the two streams by learning a spatio-temporal alignment of appearance over action features. A3M learns an attention map conditioned on the action and appearance features to better focus on the action-specific features in the frame-level appearance embeddings. This helps to improve intra-class similarity and reduce inter-class confusion for few-shot. 

Consequently, \methodname~outperforms all state-of-the-art (SoTA) unsupervised video learning methods on multiple benchmark datasets and also outperforms or performs competitively against the SoTA few-shot action recognition methods. 
To summarize, our main contributions are,
\begin{enumerate}
\item We propose \methodname~as the first unsupervised meta-learning algorithm for few-shot video action recognition.
\item \methodname~uses a two-stream network to learn action and appearance-specific features via contrastive learning over 550K unlabeled videos. It employs a novel Action-Appearance Aligned Meta-adaptation~(A3M) module that is episodically trained via hard-mined episodes to specialize for few-shot downstream tasks.
\item  \methodname~outperforms all SoTA unsupervised methods across multiple few-shot benchmarks and performs competitively to or even outperforms some of the SoTA few-shot action recognition methods.
\end{enumerate}

\section{Related Work}
\textbf{Supervised Few-shot Learning}
A typical supervised few-shot learning setting has a set of \textit{base}-classes with a large number of labeled samples and a set of \textit{novel} classes with few labeled samples (not enough for plain finetuning). It is evaluated in a \textit{meta-testing} phase where it classifies samples~(query) from the novel classes based on a few, e.g. 1 or 5 labeled examples~(support).

For images, few-shot learning approaches include metric-learning based~\cite{snell2017prototypical, vinyals2016matching, sung2018learning} that learn to minimize the distance between support and query embeddings or optimization based~\cite{finn2017model, ravi2016optimization} that develop rapidly learnable models for efficient adaptation on novel classes. Using just the base-class data inhibits generalization to novel classes. There are, therefore, approaches using data augmentation/hallucination~\cite{wang2018low,hariharan2017low} or simply training larger supervised models with larger dataset with non-episodic few-shot learning~\cite{wang2019simpleshot,chen2019closer, xian2020generalized}. There are also few-shot approaches using some form of attention/alignment module for improved performance~\cite{gidaris2018dynamic, hou2019cross, doersch2020crosstransformers} but these are image-specific and are not compatible with the action-appearance features aligned by our A3M module.

To the best of our knowledge, existing few-shot learning work for videos are all supervised approaches. ProtoGAN~\cite{kumar2019protogan} uses GANs~\cite{goodfellow2014generative} to synthesize addition examples for novel classes, CMN~\cite{zhu2018compound} uses memory augmented networks~\cite{santoro2016meta} to store video features for query matching, and R-3DFSV~\cite{xian2020generalized} uses a large pretrained 3D CNN along with weak labels to augment novel class support samples. There is also work using different forms of cross-attention/alignment such as TARN and ARN~\cite{bishay2019tarn, zhang2020few} capturing spatio-temporal dependencies via attention, OTAM~\cite{Cao_2020_CVPR} matching query-support pairs via metric-learning based temporal alignment, RVN~\cite{cao2020few} aligning support-query features via LSTMs, and AMeFu-Net~\cite{fu2020depth} aligning appearance and motion by fusing depth with RGB. Some methods also leverage auxiliary self-supervision to boost few-shot performance~\cite{gidaris2019boosting, ren2018meta, doersch2020crosstransformers, zhang2020few}. However, unlike previous formulations that either align support and query or use additional modality along with being supervised, our A3M module in \methodname~learns to align 2D and 3D features using hard-mined episodes in a purely unsupervised manner.
 

\textbf{Supervised Action Recognition}
Previous methods use either 2D CNNs with frame-level features~\cite{girdhar2017actionvlad, feichtenhofer2019slowfast} or 3D CNNs~\cite{hara2017learning,tran2015learning} with spatio-temporal features for supervised action recognition. 2D models suffer from the lack of long-term temporal reasoning while 3D models tend to overfit due to larger parameter count. To mitigate this, recent methods introduce self-attention~\cite{wang2018non}, temporal relation~\cite{zhou2018temporal}, factorized 3D convolutions~\cite{tran2018closer}, 2D replacements~\cite{xie2018rethinking}, multi-grid scheduler~\cite{wu2020multigrid} and slow-fast networks~\cite{feichtenhofer2019slowfast}. There are also two-stream networks using both 2D and 3D CNNs~\cite{wang2016temporal,carreira2017quo,tao2020rethinking} exploiting optic flow or frame residuals with RGB that we take inspiration from to design our novel action-appearance two-stream network to learn from unlabeled videos.
 
 \textbf{Unsupervised Few-Shot Learning}
Recently, unsupervised meta-learning approaches for few-shot image classification~\cite{khodadadeh2019unsupervised,  medina2020self, hsu2018unsupervised} have shown competitive performance without using base-class labels or external supervision. Our \methodname~drew inspiration from these works to leverage unlabeled videos for few-shot action recognition.


\textbf{Unsupervised Video Representation Learning}
Solving pretext tasks in images \cite{doersch2015unsupervised,zhang2016colorful,gidaris2018unsupervised, noroozi2016unsupervised} has inspired methods to learn from unlabeled videos~\cite{jing2018self, kim2019self} via pretext tasks such as sorting frames and predicting video speed~\cite{misra2016shuffle, lee2017unsupervised, xu2019self, wang2020self, benaim2020speednet, lotter2016deep, han2020memory, piergiovanni2020evolving}.
Recently, methods using contrastive learning~(InfoNCE)~\cite{oord2018representation} have been the most effective in harnessing large-scale unlabeled data~\cite{he2020momentum, chen2020simple, qian2020spatiotemporal, tao2020self,  Han20CoCLR} and perform comparably to supervised methods on vision tasks. Although these methods have shown low-shot learning capabilities, it is primarily limited to being able to finetune on \textit{in-distribution} training classes with a tiny fraction of full-labeled dataset. Unlike the proposed MetaUVFS, without any dedicated few-shot meta-learning mechanism during training, the existing unsupervised methods still require a full-size labeled dataset to optimally transfer to a downstream task with \textit{out-of-distribution} novel classes.


\begin{figure*}[t]
\begin{center}
   \includegraphics[width=\textwidth]{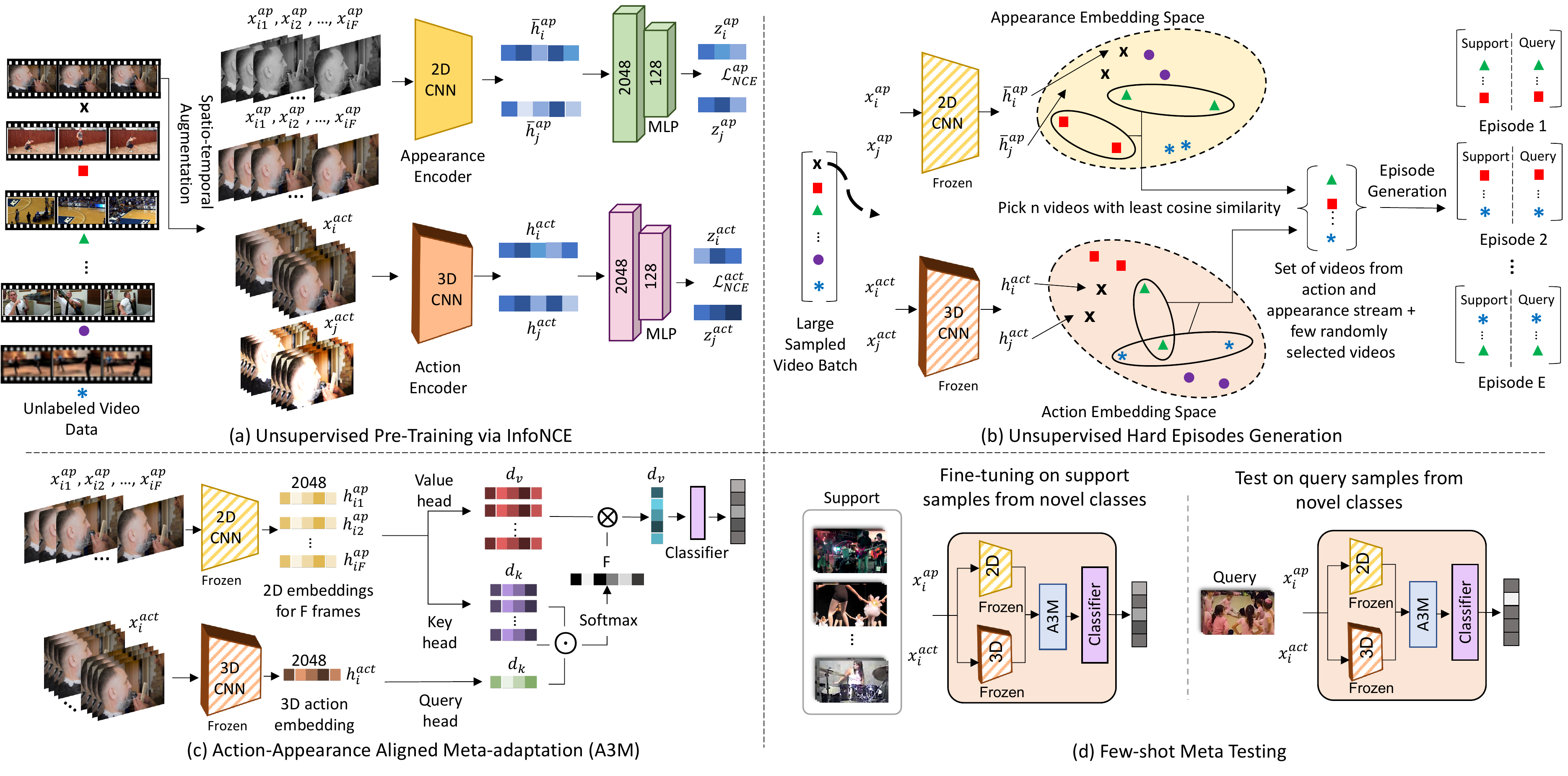}
\end{center}
\vspace{-1.5em}
   \caption{\methodname: Model Overview. (a) \textbf{Two-stream Training}: using a large unlabeled video dataset subjected to our sampling and augmentation scheme (see Section~\ref{sec:sampling}).  (b) \textbf{Hard Episode Generation} hard episodes are mined from the two-stream networks's feature space.  (c) \textbf{A3M} module learns to align appearance over action features through episodic meta-adaptation.  (d) \textbf{Meta-testing} Meta-trained A3M as a specialized few-shot classifier finetunes on novel class support to classify query videos.}
\label{fig:sesumi_model}
\vspace{-1em}
\end{figure*}


\section{Method}

We first describe the unsupervised training of the two-stream network in \methodname. We then explain the unsupervised few-shot meta-training and testing of~\methodname.


\subsection{Two-stream Video Networks}

\label{sec:twostream}
As shown in Fig.~\ref{fig:sesumi_model}a, 
~MetaUVFS has a 2D CNN-based \textit{appearance stream} $f^{ap}(\cdot)$ that captures the high-level spatial  semantics of the video. $f^{ap}(\cdot)$  encodes a sequence of $F$ frames, $\mathbf{X^{ap}} = [x^{ap}_t]_{t=1}^F$, into embeddings $\mathbf{h^{ap}}=[h^{ap}_t]_{t=1}^F$ where $h^{ap}_t=f^{ap}(x^{ap}_t)$.~$\mathbf{h^{ap}}$ are averaged to obtain $\bar{h}^{ap}$ . MetaUVFS also has a 3D CNN-based \textit{action stream} $f^{act}(\cdot)$ that captures the spatio-temporal semantics of the video. $f^{act}(\cdot)$  encodes another $F'$ frames, $\mathbf{X^{act}} = [x^{act}_t]_{t=1}^{F'}$, into a single embedding $h^{act}$, where $h^{act}=f^{act}(\mathbf{X^{act}})$. Inductive biases of using 2D and 3D convolutional kernels in the appearance and action streams respectively enable the streams to specialize in capturing the appearance and action-related video information. 


\subsection{Two-stream Unsupervised Training Objective}
The training objective of our two-streams network is based on the multi-view InfoNCE contrastive loss formulation~\cite{chen2020simple, he2020momentum, oord2018representation, tian2019contrastive} of the InfoMax principle~\cite{linsker1988self} which maximizes the mutual information between embeddings of multiple views of $x$, $x_i$ and $x_j$.
In contrastive learning, the network is trained to correctly match each input sample with an augmented version of itself among a large training batch of other samples and their respective augmentations.
We use the NT-Xent loss~\cite{chen2020simple} defined as,   
\begin{equation}
\footnotesize
    \mathcal{L}^{\text{NT-Xent}} (x_i, x_j) = - \log \frac{e^{sim(z_i, z_j)/\tau}}{\sum_{k=1}^{2N} \mathbbm{1}_{[k\neq i]} e^{sim(z_i,z_k)/\tau}}
\label{eq:nt_xent}
\end{equation}
where $sim(z_i,z_k)$ is the cosine similarity between $z_i$ and $z_k$, $\tau$ is a temperature scalar and $z_k = g(x_k)$. $N$ is the size of the mini-batch of distinct samples where each sample $x$ has $x_i$ and $x_j$ as positive augmentations. As shown in Eqn.~\ref{eq:nt_xent}, the NT-Xent loss maximizes the agreement between two augmented views $x_i$ and $x_j$ of the same input sample $x$ in a low-dimension representation space encoded by $g$.

For $x_i^{act}$ and $x_{i1}^{ap},\ldots, x_{iF}^{ap}$, the action and appearance stream encodings: $h_{i}^{act}$ and $\bar{h}_{i}^{ap}$ are fed to MLP projection heads to obtain  $z_i^{act}$ and $z_i^{ap}$.  Similarly  we obtain $z_j^{ap}$ and $z_j^{act}$ for another augmentation set $x_j^{act}$ and $x_{j1}^{ap},\ldots, x_{jF}^{ap}$. $z_i^{ap}$ and $z_j^{ap}$ are used to compute $\mathcal{L}^{ap}_{NCE}$ contrastive loss to train the appearance encoder, while $\mathcal{L}^{act}_{NCE}$ is computed using $z_i^{act}$ and $z_j^{act}$ to train the action encoder. 

\subsection{Unsupervised Meta-learning for Video Few-Shot~(MetaUVFS)}
\methodname~explicitly trains a few-shot meta-learner via episodic training to improve performance on the downstream few-shot tasks having novel classes. \methodname~first generates episodes at video instance level using noise-contrastive embeddings without any supervision and imposes a hardness threshold to boost few-shot meta-learning. Using these generated episodes, \methodname~trains a novel Action-Appearance Aligned Meta-adaptation~(A3M) module to align and relate action and appearance features, and output an embedding that can more effectively generalize to novel classes in few-shot testing. The episodic training of \methodname~imposes a downstream task-specific prior on the unsupervised model features that reduces the gap between train and test settings, thereby improving performance.

\subsubsection{Unsupervised Hard Episodes Generation}
To simulate the meta-testing episodic setting during training, we leverage the unlabeled video data to generate meaningful episodes for meta-training the A3M module.
We generate 1-shot, 5-way classification episodes (similar to the downstream few-shot task) where the support and query for each class are formed using spatio-temporal augmentations (Sec.~\ref{sec:sampling}) of an unlabeled video sample. In this way, the classification happens at the instance level (\ie each video behaves as its own class) and the task is to classify a query augmentation belonging to the correct video sample.  A simple approach would be to randomly sample unlabeled videos and process their augmentations into episodes. 
However, the InfoNCE contrastive learning pushes the embeddings, $h_{i}^{act}$ and $\bar{h}_{i}^{ap}$, for the augmentations of a video $x_i$ already very close to each other compared to embeddings for augmentations for other videos. Thus, randomly sampled videos will provide  A3M module episodes that can be trivially solved and will impede any meaningful learning. As shown in Fig.~\ref{fig:sesumi_model}b, to incentivize learning and generate meaningful episodes, we mine episodes where we select hard video instances whose augmentations lie far from each other in the embedding space of the trained two-stream encoders. We feedforward a large batch of video augmentations through the trained and frozen action and appearance encoders. For each encoder, we select $n$ videos which have the lowest cosine similarity among its augmentations. We pool the set of the videos collected this way from the two encoders and extend this video set by another $10\%$ with randomly sampled videos for exploration and to cover all video samples on expectation. We then sample $E$ episodes from this set of videos for one training iteration of few-shot training. Selecting $n$ videos for both action and appearance enables the few-shot meta-learner to reduce confusion from both action and appearance perspectives~(Fig.~\ref{fig:teaser}).

\subsubsection{A3M: Action-Appearance~Aligned~Meta-adaptation}
As shown in Figure~\ref{fig:teaser}, it is important for the model to attend to both action and appearance-related aspects of a video in correspondence to each other to enhance intra-class relationship and avoid inter-class confusion, particularly when learning from very few labeled samples. To this end, we design a novel cross-attention module for action-appearance aligned meta-adaptation, A3M, that is trained using episodic few-shot learning to meta-learn to cross-align action with appearance-related features. 

The A3M module learns to establish a soft correspondence between the action and appearance features using attention-based Transformers~\cite{vaswani2017attention}. As shown in Fig.~\ref{fig:sesumi_model}c, we parameterize three linear mappings, key-head $\mathbf{K}: \mathbb{R}^D \rightarrow \mathbb{R}^{d_k}$, value-head $\mathbf{V}: \mathbb{R}^D \rightarrow \mathbb{R}^{d_v}$ and query-head $\mathbf{Q}: \mathbb{R}^D \rightarrow \mathbb{R}^{d_k}$ for this purpose where $d_k$ and $d_v$ are the size of the key and value embeddings, respectively. 
We generate key-value pairs using $\mathbf{K}$ and $\mathbf{V}$ for the frame-level representations, $h_{i1}^{ap},\ldots, h_{iF}^{ap}$, from the 2D appearance encoder. Let $\mathbf{k}_m = \mathbf{K} \cdot h_m^{ap}$ and $\mathbf{v}_m = \mathbf{V} \cdot h_m^{ap}$ form the key-value pair for the $m^{th}$ frame-level representation for unlabeled $x_i$. We also generate a query embedding, $q = \mathbf{Q} \cdot h^{act}$, for the spatio-temporal feature, $h^{act}$, from the 3D action encoder using $\mathbf{Q}$. We then compute the dot-product attention scores between the keys and the query, and normalize the scores via softmax over all key embeddings as, 
\begin{equation}
\footnotesize
a_m = \frac{\exp(\mathbf{k}_m \cdot \mathbf{q}) / \sqrt{d_k}}{\sum_{t} \exp(\mathbf{k}_t \cdot \mathbf{q} / \sqrt{d_k})}
\label{eq:attn_ma2m}
\end{equation}
where $a_m$ is the attention score for the $m^{th}$ frame embedding. These attention scores provide a soft correspondence that align and relate the action information with the appearance of the video.
The attention scores are then combined with the value head embeddings and aggregated via sum to obtain a single feature embedding, $\mathbf{h^{A3M}} = \sum_{m} a_m \mathbf{v}_m$. As the attention scores are computed via a combination of action and appearance features, they weigh the appearance features to focus on the most action-relevant parts. The aggregated embedding $\mathbf{h^{A3M}}$, conditioned on both action and appearance information, is therefore better equipped than naive concatenation for few-shot tasks. 



\subsubsection{Few-Shot Meta-Training}
We leverage Model-Agnostic Meta Learning~(MAML)~\cite{finn2017model} to train the network to learn to adapt to a new task of novel action classes with few labeled samples. Once we train the action and appearance streams, we freeze the two backbones and train $f_\theta$ comprising of the A3M module along with a classifier layer during the few-shot episodic meta-training. The action-appearance aligned feature embedding from the A3M module is $l_2$-normalized before being fed to the classifier. For each generated episode $e \in E$ in a training iteration, we generate $s$ support augmentations for sampled videos and compute adapted parameters with gradient descent of the cross-entropy classification loss $\mathcal{L}$ over $f_\theta$ as $\theta'_e  = \theta - \alpha \nabla_\theta \mathcal{L}_e (f_\theta)$ where $\alpha$ is the adaptation learning rate. We then generate $q$ query augmentations for videos in episode $e$ to compute the loss $\mathcal{L}$ using adapted parameters $\theta'_e$ as $\mathcal{L}_e (f_{\theta'_e})$. We repeat this for all $E$ episodes and finally update $\theta$ at the end of the training iteration as $\theta \leftarrow \theta - \beta \nabla_\theta \sum_e^E \mathcal{L}_e (f_{\theta'_e})$
where $\beta$ is the learning rate for the meta-learner optimizer.
\subsubsection{Few-Shot Meta-Testing}
\label{sec:fs_testing}
Once trained, we test 
\methodname~
by finetuning on multiple few-shot test episodes. As can be seen in Fig.~\ref{fig:sesumi_model}d, for each episode, we freeze the action-appearance encoders and finetune the A3M and classifier layers which has been meta-trained. After every episode, we refresh the parameters of A3M and classifier layers for the next episode.



\section{Experiments and Results}
\begin{table*}[t]
\centering
\resizebox{\textwidth}{!}{%
\begin{tabular}{|l|c|c|c|c|c|c|c|c|}
\hline
\multirow{2}{*}{Methods} & \multicolumn{2}{c|}{Supervision} & \multicolumn{2}{c|}{UCF101} & \multicolumn{2}{c|}{HMDB51} & \multicolumn{2}{c|}{Kinetics100} \\ \cline{2-9} 
 & \multicolumn{1}{l|}{Pretraining} & \multicolumn{1}{l|}{Base-Class} & 1-shot & 5-shot & 1-shot & 5-shot & 1-shot & 5-shot \\ \hline
Matching Net \cite{zhu2018compound} & Imagenet-2D & Yes & - & - & - & - & 53.3 & 74.6 \\ 
MAML \cite{zhu2018compound} & Imagenet-2D & Yes & - & - & - & - & 54.2 & 75.3 \\ 
CMN \cite{zhu2018compound} & Imagenet-2D & Yes & - & - & - & - & 60.5 & 78.9 \\ 
TARN \cite{bishay2019tarn} & Sports-1M & Yes & - & - & - & - & 66.6 & 80.7 \\ 
OTAM \cite{Cao_2020_CVPR} & Imagenet-2D & Yes & - & - & - & - & 73.0 & 85.8 \\
R-3DFSV  \cite{xian2020generalized} & Sports-1M & Yes & - & - & - & - & \textcolor{blue}{95.3} & \textcolor{blue}{97.8} \\ 
ProtoGAN \cite{kumar2019protogan} & Sports-1M  & Yes  & 57.8  $\pm$  3.0 & 80.2  $\pm$  1.3 & 34.7  $\pm$  9.20 & 54.0  $\pm$  3.90 & - & - \\ 
AmeFu-Net \cite{fu2020depth} & Imagenet-2D & Yes & 85.1 & 95.5 & 60.2 & 75.5  & 74.1  & 86.8 \\ 
RVN \cite{cao2020few} & Kinetics-400 & Yes  & \textcolor{blue}{88.71  $\pm$  0.19} & \textcolor{blue}{96.78  $\pm$  0.08} & \textcolor{blue}{63.43  $\pm$  0.28} & \textcolor{blue}{79.69  $\pm$  0.20} & - & - \\ \hline
ARN  \cite{zhang2020few}  & No & Yes & 66.32  $\pm$  0.99 & 83.12  $\pm$  0.70 & 45.15  $\pm$  0.96 & 60.56  $\pm$  0.86 & 63.7 & 82.4 \\ \hline

3DRotNet \cite{jing2018self} & No & No & 39.43 $\pm$ 0.48	& 33.61 $\pm$ 0.34 & 32.35 $\pm$  0.42 & 27.84 $\pm$ 0.40 & 27.53 $\pm$  0.36 &	25.54 $\pm$ 0.39 \\ 
VCOP \cite{xu2019self}  & No & No & 32.91 $\pm$  0.42 & 39.11 $\pm$ 0.37 & 27.80  $\pm$  0.37 & 31.56 $\pm$ 0.35 & 26.48 $\pm$ 0.37  & 28.87 $\pm$ 0.36 \\ 
IIC \cite{tao2020self} & No & No & 56.81 $\pm$ 0.46 & 78.74  $\pm$  0.37 & 34.66 $\pm$  0.41 & 49.57 $\pm$  0.44 & 37.73 $\pm$ 0.43 & 51.11 $\pm$ 0.43 \\ 
Pace Prediction \cite{wang2020self} & No & No &  25.58 $\pm$ 0.33 & 26.58 $\pm$ 0.31 & 26.21 $\pm$ 0.33 & 27.09 $\pm$ 0.31 & 22.42 $\pm$ 0.33 & 22.94  $\pm$ 0.30 \\ 
MemDPC \cite{Han20} & No & No & 49.27 $\pm$ 0.44 & 67.38 $\pm$ 0.45 & 30.33 $\pm$ 0.40 & 41.15 $\pm$ 0.42 & 42.01 $\pm$ 0.41 & 53.90 $\pm$ 0.43 \\ 
CoCLR  \cite{Han20CoCLR} & No & No & 51.99 $\pm$ 0.46 & 72.17 $\pm$ 0.42 & 31.29 $\pm$ 0.40 & 44.92 $\pm$ 0.45 & 37.59 $\pm$ 0.42 & 51.11 $\pm$ 0.43 \\ 

CVRL \cite{qian2020spatiotemporal} & No & No & 63.00  $\pm$  0.41 & 87.80  $\pm$  0.30 & 44.21  $\pm$  0.45 & 60.35  $\pm$  0.45 & 53.26  $\pm$  0.48 & 71.39  $\pm$  0.44\\ 
 \hline


\methodname~(Ours)  & No & No & \textbf{76.38 $\pm$ 0.40}  & \textbf{92.50 $\pm$ 0.24} & \textbf{47.55 $\pm$ 0.45}  &  \textbf{66.13 $\pm$ 0.33} & \textbf{62.80 $\pm$ 0.45} & \textbf{79.55 $\pm$ 0.39} \\ 
\hline

\end{tabular}%
}
\caption{Results on UCF101, HMDB51 and Kinetics100 datasets for 5-way, 1-shot and 5-shot few-shot action recogniton. Our method \methodname~ outperforms SoTA methods on unsupervised video representations by large margins on few-shot benchmarks. We also show competitive performance  \wrt supervised few-shot video approaches. Moreover, on UCF101 and HMDB51, \methodname~is able to outperform ARN that uses only base-class supervision. \methodname~also outperforms ProtoGAN on UCF101 and HMDB51, and CMN on Kinetics100. Values in \textcolor{blue}{blue} represent SoTA across all levels of supervision.}
\label{tab:fsvar}
\vspace{-1em}
\end{table*}
\subsection{Datasets} 
We evaluate \methodname~on three publicly-available few-shot datasets: Kinetics100 \cite{carreira2019short, zhu2018compound}, UCF101 \cite{soomro2012ucf101} and HMDB51 \cite{kuehne2011hmdb}. Following~\cite{zhu2018compound}, we obtain the \textit{few-shot}  train/validation/test splits with 64/12/24 non-overlapping classes for Kinetics100. For UCF101 and HMDB51, we follow the \textit{few-shot} split from \cite{zhang2020few}. UCF101 contains 100 classes split into 70/10/20 and HMDB51 contains 51 classes split as 31/10/10. The test splits of each dataset are used for novel class evaluation in the meta-testing phase. For the unsupervised training of \methodname's two-stream networks, we leverage Kinetics700~\cite{carreira2019short} without using any labels. Kinetics700 is a large-scale video classification dataset that covers 700 human action classes including human-object and human-human interactions. To increase the size of our unlabeled training data, we also include the videos from the base-classes of Kinetics100, UCF101, and HMDB51, without the labels. Altogether, we obtain around $550$K video clips with a duration of around 10s each (25 FPS). We take extra precaution to ensure that there is no video in the training dataset belonging to the union of all the novel classes across all three evaluation datasets. This is to ensure that our testing is truly on a disjoint set of unseen classes.               

\subsection{Implementation Details}
\noindent{\bf Data Sampling and Augmentation}
\label{sec:sampling}
We develop a spatio-temporal sampling protocol that is most optimal for the unsupervised two-stream training and A3M-based few-shot training/testing. For an input video, the 2D appearance stream encodes 8 input frames where 1 frame is randomly sampled from each of 8 segments equally-partitioned along the video length. With focus on spatial information, we use a higher frame resolution of $224 \times 224$. We refer to this as $8\times1$.  For the 3D-action stream, with the goal of encoding fine-grained spatio-temporal action information across video segments, we sample 4 clips across 4 equidistant segments of the video to form a 16 frame input. To balance the spatio-temporal information, we use a lower frame resolution of $112 \times 112$. We refer to this as $4\times4$. 
We follow SimCLR's protocol for spatial augmentation~\cite{chen2020simple}: a composition of Random crops, Random horizontal flips, Random Color Jitter, Random grayscale, Gaussian blur. The spatial augmentation is clip-wise consistent, \ie, the random seed is fixed across all frames of a video augmentation~\cite{xu2019self,han2020memory}.  
\paragraph{MetaUVFS Training} 
We use ResNet50~\cite{he2016deep} backbone to train the 2D appearance stream and its 3D counterpart, ResNet50-3D~\cite{hara3dcnns}, for the 3D action stream. The dimension of $z^{ap}$ and $z^{act}$ obtained from the MLP projection head is 128~(similar to~\cite{chen2020simple}).
 We first train the action and appearance streams individually using losses $\mathcal{L}^{ap}_{NCE}$ and $\mathcal{L}^{act}_{NCE}$ respectively. We use a batch size of 512 and train both models for 300 epochs on 64 NVIDIA P100 GPUs. Following~\cite{he2019bag, goyal2017accurate}, we do a gradual learning rate (LR) warmup for 5 epochs followed by a half-period cosine learning rate decay with SGD optimizer and 0.9 momentum. With 0.001 per-gpu LR, we also linearly scale the LR to 0.064. 

For hard-mining episodes, $n$ is set to 32. We set $d_k=128$ and $d_v=2048$ for the A3M module. For MAML, we set $E=10$, $\alpha=0.001$ and $\beta=10$. We train for $20,000$ iterations using Adam optimizer and cosine annealing~\cite{antoniou2018train} for a total of $200K$ unsupervised hard-mined episodes. For more details, please refer to the supplementary material.
 
\paragraph{Few-shot Evaluation} We evaluate \methodname~on all three datasets based on 5-way, 1-shot and 5-way, 5-shot settings as is standard in few-shot learning literature.
For each episode, 5 classes are randomly sampled from the set of novel classes for classification and training happens on 1 and 5 support samples per class respectively. In all settings, Top-1 accuracy is reported on 1 query sample per class. 
In each experiment, we randomly sample 10,000 episodes for few-shot meta-testing and report the average accuracy at the 95\% confidence interval. Finetuning is done at a constant learning rate of 10 for 50 epochs for all experiments.


\begin{table*}[t]
\centering
\resizebox{\textwidth}{!}{%
\begin{tabular}{|cccc|c|c|c|c|c|c|}
\hline
\multirow{2}{*}{Action} & 
\multirow{2}{*}{Appearance} &
\multirow{2}{*}{A3M} &
\multirow{2}{*}{Hard Episodes} &
\multicolumn{2}{c|}{UCF101} & \multicolumn{2}{c|}{HMDB51} & \multicolumn{2}{c|}{Kinetics100} \\ \cline{5-10} 
 & & & & 1-shot & 5-shot & 1-shot & 5-shot & 1-shot & 5-shot \\ \hline

 \checkmark & & & \checkmark & 66.97 $\pm$ 0.44 & 88.64 $\pm$ 0.30  & 44.56 $\pm$ 0.45   &  61.03 $\pm$ 0.45   & 53.56 $\pm$ 0.48 & 71.31 $\pm$ 0.44 \\ 
  & \checkmark & & \checkmark & 66.10 $\pm$ 0.45 & 84.58 $\pm$ 0.34  & 39.97 $\pm$  0.46   &  58.20 $\pm$ 0.46   & 50.91 $\pm$ 0.48 & 69.47 $\pm$ 0.46 \\ 
\checkmark  &  \checkmark   & &  & 71.82 $\pm$ 0.42  & 89.93  $\pm$ 0.27  & 44.64 $\pm$ 0.46    & 62.40 $\pm$ 0.45   & 57.52 $\pm$ 0.46 & 75.21 $\pm$ 0.41  \\ 
 \checkmark  & \checkmark & & \checkmark & 73.02 $\pm$ 0.42   &  91.38 $\pm$  0.26 & 44.89  $\pm$ 0.46   &   64.96 $\pm$ 0.44   &  59.16 $\pm$ 0.44   & 77.42  $\pm$ 0.40   \\ 
\checkmark  & \checkmark & \checkmark &  & 73.97 $\pm$ 0.41 & 91.50 $\pm$ 0.26 & 45.84 $\pm$ 0.45 & 64.68 $\pm$ 0.41 & 59.88 $\pm$ 0.45  & 77.64 $\pm$ 0.39 \\ 
 \checkmark & \checkmark  & \checkmark  & \checkmark    & \textbf{76.38 $\pm$ 0.40}  & \textbf{92.50 $\pm$ 0.24} & \textbf{47.55 $\pm$ 0.45}  &  \textbf{66.13 $\pm$ 0.33} & \textbf{62.80 $\pm$ 0.45} & \textbf{79.55 $\pm$ 0.39} \\ 
\hline

\end{tabular}%
}
\caption{Ablation study of \methodname~ highlights the superior few-shot performance of meta-training two-stream feature representations over individual action and appearance streams. The performance is further boosted by action-appearance feature alignment by the A3M module. Moreover, mining unsupervised hard episodes is crucial for effectively training the A3M module.}
\label{tab:ablation}
\vspace{-1em}
\end{table*}
\subsection{Compare to SoTA Unsupervised Approaches}
Table~\ref{tab:fsvar} compares \methodname~with various state-of-the-art supervised and unsupervised methods on different few-shot settings and datasets. We categorize the different techniques based on the amount of supervision in terms of base-class data (`Yes' in \textit{Base-Class}) and surrogate supervision, \ie, initializing the network using the weights pretrained on a large-scale supervised image/video data (`Yes' in \textit{Pretrained Weights}). Cells are left blank if there are no publicly available results for that setting. 

The second part of Table~\ref{tab:fsvar} compares \methodname~with various state-of-the-art methods that leverage unlabeled videos for representation learning. To the best of our knowledge, \methodname~is the first approach that specializes in few-shot action recognition in a purely unsupervised manner. Hence, there is no publicly available benchmark for the performance of existing video-based unsupervised techniques on few-shot action recognition. We took the initiative to assess these approaches on our few-shot test-bed using the same hyperparameters for few-shot meta-testing as \methodname. Many of these approaches are originally trained on a relatively small unlabeled dataset. Therefore, for a fair comparison, we train these methods on our large-scale unlabeled dataset using their publicly available code.

As shown in Table~\ref{tab:fsvar}, \methodname~is able to clearly outperform all state-of-the-art unsupervised methods on the task of few-shot action recognition by at least 13.38\%, 3.34\% and 9.54\% (absolute increase) on UCF101, HMDB51 and Kinetics100 1-shot, 5-way benchmark respectively. Among the methods we compare, IIC~\cite{tao2020self}, CVRL~\cite{qian2020spatiotemporal} and CoCLR~(RGB only)~\cite{Han20CoCLR} also use contrastive loss for unsupervised training. The superior performance of \methodname~in comparison to these methods indicate that our approach of jointly leveraging and aligning action and appearance along with meta-training episodically for few-shot plays an integral role in performing effectively when the downstream task lies in the low-shot regime.

\subsection{Compare to SoTA Supervised Few-shot Works}
The first part of Table~\ref{tab:fsvar} compares \methodname~with various state-of-the-art supervised few-shot action recognition methods.
We can observe that compared to ARN~\cite{zhang2020few} that uses only base-class data as supervision, \methodname~significantly outperforms on UCF101 and HMDB51, and performs competitively on Kinetics100. Furthermore, \methodname~is even able to outperform some of the supervised methods that use both pretrained weights and base-class labels for supervision such as ProtoGAN~\cite{kumar2019protogan} on UCF101 and HMDB51, and CMN~\cite{zhu2018compound} on Kinetics100. It is worth noting that, unlike these methods that need to train separate models to obtain results on the different datasets, \methodname~trains a \textit{single} unsupervised model to achieve all results. This single model either outperforms or performs competitively compared to supervised methods across all three datasets.

\subsection{\methodname: Ablation Study}
We conduct an ablation study where we isolate individual aspects of \methodname~and quantify their impact on the few-shot performance. Table~\ref{tab:ablation} summarizes the results. We train all ablation experiments using MAML as the few-shot algorithm. 

We first conduct experiments without the A3M module where the network consists of only the action stream, only the appearance stream and dual action-appearance stream (Table~\ref{tab:ablation}, Rows 1, 2, 4). Without the A3M module, for the one-stream setting, we directly feed the features from the available stream (averaging appearance features over 8 frames) to the classifier layer for few-shot episodic meta-training and later for meta-testing; for the two-stream setting, we simply concatenate the action features and appearance features (averaged over 8 frames) and feed them to the classifier for few-shot episodic meta-training. All three experiments use unsupervised hard-mined episodes. In the absence of either the action or the appearance stream, only the features of the available stream are used to mine episodes. We can observe from Table~\ref{tab:ablation}~(Rows 1,~2,~4) that the few-shot performance is significantly worse when either action or appearance stream is missing compared to when both are present. This is because when only a few support samples are available to learn for a set of novel classes, the likelihood of the model to make mistakes reduces sharply in the presence of both streams as it allows the network more ways to activate and respond to the representative features necessary for correct classification. We can also compare Row 1~(Action stream only) with CVRL~\cite{qian2020spatiotemporal} in Table~\ref{tab:fsvar}. CVRL backbone is similar to our 3D action stream. However, due to an explicit few-shot training phase, our \textit{Action only} baseline performs consistently better than CVRL.

Rows 4 and 6 in Table~\ref{tab:ablation} highlight the impact of the A3M module by aligning action-appearance as part of few-shot training. Our proposed A3M module in MetaUVFS results in an average absolute improvement of 3.22\% and 1.47\% on 5-way, 1-shot and 5-way, 5-shot benchmarks across all datasets. Aligning the action and appearance features during few-shot episodic training significantly improves the model’s ability to \textit{attend} to the most representative video aspects while leveraging the inductive biases of both 2D and 3D CNNs to learn complementary representation that boosts few-shot performance.  

We then perform an ablation where we train our method episodically without mining hard episodes based on noise-contrastive embeddings (Table~\ref{tab:ablation}, Row 5). Comparing Rows 5 and 6, we can observe a significant reduction in performance without hard episodes, underlining the importance of mining hard episodes to the few-shot episodic training of \methodname. This is because in the absence of hard episodes, the randomly sampled videos in a training episode are such that the action and appearance embeddings fed for support and query augmentation samples to the A3M module during training are already easily separable. This severely compromises the training of A3M and makes it behave close to an identity function, as evident from Row 5's only slightly higher performance than Row 4 where A3M is not present. 

We additionally perform an experiment where both A3M and hard episodes are absent during training (Table~\ref{tab:ablation}, Row 3). We can observe that this setting results in a statistically significant reduction in performance compared to when  A3M and/or hard episodes are employed for training (Rows 4-6).

\begin{table*}[t]
\footnotesize
\centering
\begin{tabular}{l|cccccc}
\cline{1-7}
 & \multicolumn{3}{c|}{Appearance only (224$\times$224)} & \multicolumn{3}{c}{Action only  (112$\times$112)} \\ \hline
Sampling & 4$\times$1 & 8$\times$1 & \multicolumn{1}{c|}{16$\times$1} & 32$\xrightarrow{}$16 & 4$\times$4 & 8$\xrightarrow{}$4$\times$4 \\ \hline
\multicolumn{1}{c|}{1-shot} & 50.54 $\pm$ 0.48 & 50.91 $\pm$ 0.48 & \multicolumn{1}{c|}{51.0 $\pm$ 0.42} & 50.26 $\pm$0.52 & 53.56 $\pm$ 0.48 & 53.09  $\pm$ 0.42 \\ \hline
 & \multicolumn{6}{c}{\methodname (Appearance + Action + A3M, ours)} \\ \hline
Sampling & 4$\times$1, 4$\times$4 & 4$\times$1, 8$\xrightarrow{}$4$\times$4 & 8$\times$1, 4$\times$4 & 8$\times$1, 8$\xrightarrow{}$4$\times$4 & 16$\times$1, 4$\times$4 & 16$\times$1, 8$\xrightarrow{}$4$\times$4 \\ \hline
\multicolumn{1}{c|}{1-shot} & 60.76 $\pm$ 0.44 & 60.69 $\pm$ 0.45 & \textbf{62.80 $\pm$ 0.45} & 61.34 $\pm$0.48 & 60.30 $\pm$0.41 & 60.91  $\pm$ 0.47 \\ \hline
\end{tabular}%
\caption{Comparing sampling strategies evaluated on Kinetics100 for 1-shot, 5-way. $\xrightarrow{}$ denotes downsampling from A to B value.}
\label{tab:Sampling}
\vspace{-2em}
\end{table*}
\subsection{Discussion}
\label{sec:discussion}
{\noindent \bf Impact of Frame Sampling.} Since the two-streams in \methodname~specialize both in terms of architecture and their function, we observe that the sampling strategy in choosing the frames as input to both streams along with their frame resolution make a difference in the performance. Table~\ref{tab:Sampling} provides an analysis of the few-shot performance for 5-way, 1-shot settings on Kinetics100 across different sampling strategies for both 3D action and 2D appearance streams first individually and then in combination. We observe that for the 3D stream, choosing a $4 \times 4$ sampling, \ie, sampling 4 segments of 4 frames uniformly over the entire video length provides a $3.3\%$ improvement over sampling 16 frames from 32 consecutive frames with a stride of 2. Similarly, for the 2D stream, we find that $16 \times 1$ and $8 \times 1$ sampling, \ie, sampling 1 frame from 16 or 8 segments over the entire video length as  most effective. In our two-stream setting, we find $8 \times 1$, $4 \times 4$ as the optimal sampling scheme. 

\begin{figure}
\begin{center}
  \includegraphics[width=\columnwidth]{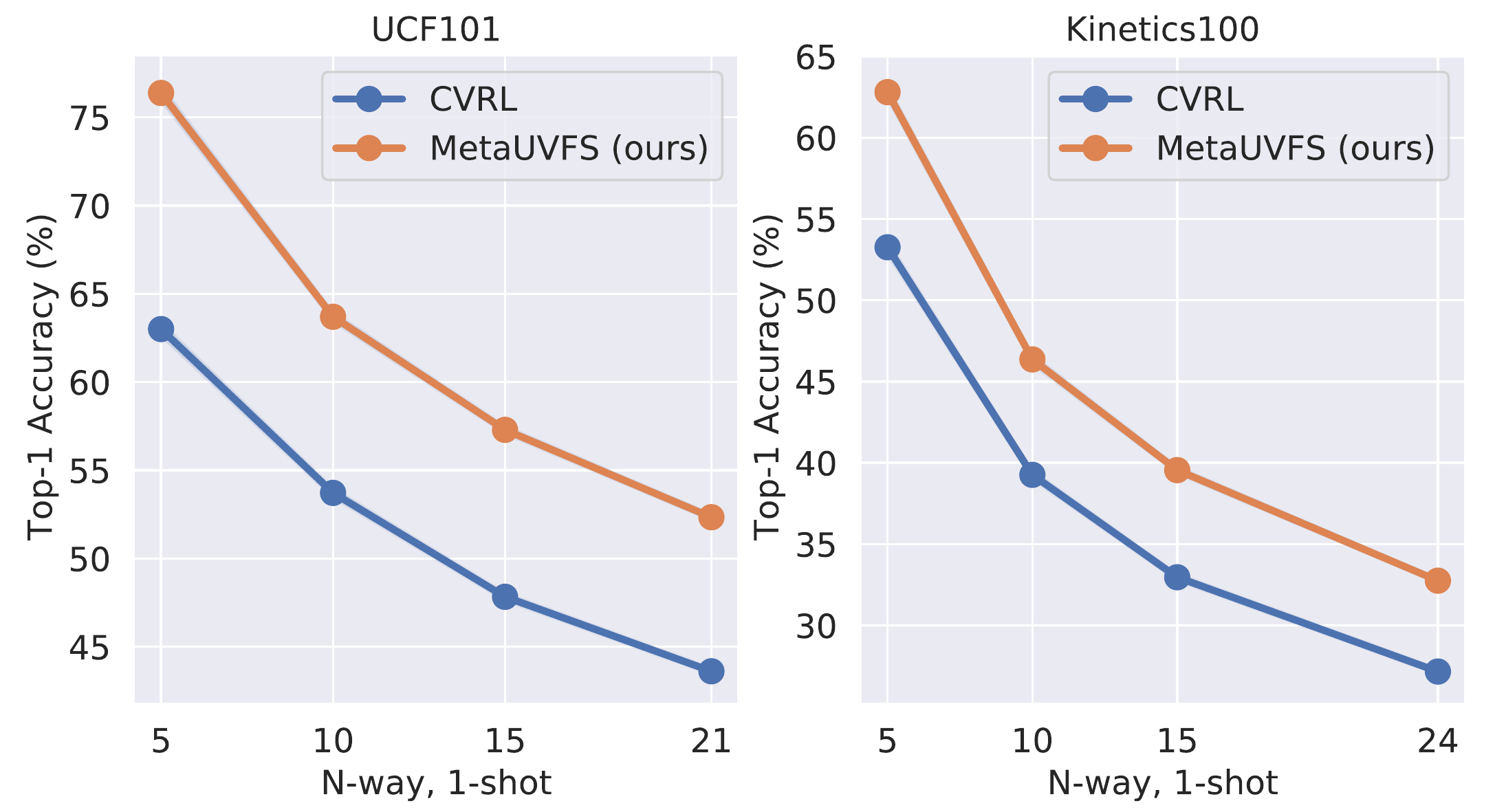}
\end{center}
\vspace{-1em}
  \caption{
  Comparison of \methodname~ with CVRL~\cite{qian2020spatiotemporal} on UCF101 and Kinetics100 dataset on 1-shot few-shot for N-way classification where N ranges from 5 to over 20.}
\label{fig:many_way}
\vspace{-1em}
\end{figure}

\begin{table}[t]
\centering
\resizebox{0.8\columnwidth}{!}{%
\begin{tabular}{l|c|c}
\hline
Few-shot Algorithm & 1-shot & 5-shot \\ \hline
ProtoNet~\cite{snell2017prototypical} & 31.20  $\pm$  0.39 & 55.96  $\pm$  0.44 \\
Baseline++~\cite{chen2018closer} & 56.10  $\pm$  0.47 & 73.37  $\pm$  0.43 \\
ProtoMAML~\cite{triantafillou2019meta} & 62.13 $\pm$ 0.46 & 78.65 $\pm$ 0.40  \\
MAML~\cite{finn2017model} & \textbf{62.80 $\pm$ 0.45} & \textbf{79.55 $\pm$ 0.39} \\ \hline
\end{tabular}
}
\caption{Comparison between different few-shot meta-learning algorithms on Kinetics100 5-way, 1/5-shot dataset.}
\vspace{-1.5em}
\label{tab:meta_learning}
\end{table}
{\noindent \bf Meta-learning Algorithm.} We further validate the choice of MAML as our few-shot meta-learning algorithm by assessing other popular few-shot approaches in the literature. To compare with ProtoNet~\cite{snell2017prototypical} and ProtoMAML~\cite{triantafillou2019meta}, we repurpose the output of A3M module to compute prototypes across support samples (augmentations) and compare against query samples to compute the loss during training. For few-shot testing on novel classes, ProtoNet matches the query samples with prototypes that are computed from support samples to assign class label based on the best matched prototype. For ProtoMAML, we reshape the prototypes computed from support samples as parameters of the classifier layer which is finetuned along with the A3M module as per Sec.~\ref{sec:fs_testing}. We also compare with Baseline++~\cite{chen2018closer} where we use meta-trained parameters for A3M and classifier but change the few-shot finetuning during testing to Baseline++. As shown in Table~\ref{tab:meta_learning}, we find that our few-shot meta-testing protocol of using MAML with $l_2$-normalized embedding as input to the classifier significantly outperforms other few-shot learning methods. We also observe that, in general, few-shot methods employing finetuning during few-shot testing tend to perform better. We believe this is due to extra adaptation steps needed during testing because of the absence of supervision during training. 
\begin{table}[h]
\centering
\resizebox{0.9\columnwidth}{!}{%
\begin{tabular}{l|c|c}
\hline
Streams & 1-shot & 5-shot \\ \hline
Appearance + Appearance & 55.75  $\pm$  0.46 & 72.23  $\pm$  0.43 \\
Action + Action & 54.25  $\pm$  0.47 & 73.21  $\pm$  0.42 \\
Action + Appearance & \textbf{59.16 $\pm$ 0.44}   & \textbf{77.42  $\pm$ 0.40}    \\ \hline
\end{tabular}
}
\caption{Comparison of using Action and Appearance streams in \methodname~with using two Action or two Appearance streams. Results are without A3M module for fair comparison.}
\label{table:aa}
\vspace{-1em}
\end{table}
{\noindent \bf Significance of Action-Appearance} We conduct an experiment where both the streams are either 3D CNNs~(action) or 2D CNNs~(appearance).  This delineates the impact of having complementary action and appearance streams on few-shot performance from the impact of increase in the parameter count due to an additional backbone. We train them using MAML with hard episodes without A3M for fair comparison. We can observe from Table~\ref{table:aa} that although having more parameters with either two appearance or two action streams improves the performance compared to single stream, the improvement is significantly higher when a combination of action and appearance streams is used that helps to leverage more diverse 2D/3D representations to learn more generalizable few-shot video representations.  

{\noindent \bf Many-Way Few Shot.} We go beyond the 5-way 1-shot setting by increasing the number of novel classes 
to evaluate \methodname~on a more challenging and \textit{in-the-wild} few-shot many-way classification task on UCF101 and Kinetics100. Fig.~\ref{fig:many_way} shows a plot for this experiment. Although, as expected, the performance reduces with increasing N, \methodname~is still able to outperform the best-performing unsupervised baseline method, CVRL, by a significant margin for all N-way 1-shot classification tasks considered.
This proves that \methodname~is more robust even in extreme few-shot scenarios where the inter-class confusion is higher. 



\section{Conclusion}
We propose a novel unsupervised meta-learning algorithm, \methodname, for few-shot video action recognition. It leverages large-scale unlabeled video data to learn unsupervised video features from a two-stream action-appearance network. It further performs explicit few-shot episodic meta-learning over unsupervised hard-mined episodes using a novel Action-Appearance Aligned Meta-adaptation~(A3M) module. The A3M module learns to align the 3D action with 2D appearance features to learn an embedding that is more effective in focusing on the action-specific features of a video for the few-shot downstream task. Through extensive experiments, we demonstrate that using an explicit few-shot learner and action-appearance aligned features makes \methodname~significantly better suited for downstream few-shot tasks compared to all state-of-the-art unsupervised methods. Moreover, \methodname~performs competitively and sometimes even outperforms SoTA supervised few-shot methods. 

\section*{Acknowledgement}
At Oregon State University, Jay Patravali is supported by DARPA grant N66001-19-2-4035.

{\small
\bibliographystyle{ieee_fullname}
\bibliography{egbib}
}

\end{document}